\pdfoutput=1

\documentclass[11pt]{article}

\usepackage[table,dvipsnames]{xcolor}
\usepackage{ACL2023}

\usepackage{times}
\usepackage{latexsym}

\usepackage{caption}
\usepackage{subcaption}
\usepackage{graphicx}
\usepackage{verbatim}
\usepackage{enumitem}
\usepackage{booktabs}
\usepackage{amsfonts}
\usepackage{tabularx}

\usepackage{pifont}
\usepackage[T1]{fontenc}

\usepackage[utf8]{inputenc}

\usepackage{csquotes}
\usepackage{microtype}

\usepackage{inconsolata}

\title{The crime of being poor}

  \author{Georgina Curto,$^1$  Svetlana Kiritchenko,$^2$ Isar Nejadgholi,$^2$ and Kathleen C. Fraser$^2$ \\
  $^1$University of Notre Dame, Notre Dame, USA \\
  $^2$National Research Council Canada, Ottawa, Canada \\
 \footnotesize \texttt{gcurtore@nd.edu, \{Svetlana.Kiritchenko,Isar.Nejadgholi,Kathleen.Fraser\}@nrc-cnrc.gc.ca}\\
 }

\begin{document}
\maketitle

\begin{abstract}
The criminalization of poverty has been widely denounced as a collective bias against the most vulnerable. NGOs and international organizations claim that the poor are blamed for their situation, are more often associated with criminal offenses than the wealthy strata of society and even incur criminal offenses simply as a result of being poor. While no evidence has been found in the literature that correlates poverty and overall criminality rates, this paper offers evidence of a collective belief that associates both concepts. This brief report measures the societal bias that correlates criminality with the poor, as compared to the rich, by using Natural Language Processing (NLP) techniques in Twitter. The paper quantifies the level of crime-poverty bias in a panel of eight different English-speaking countries. The regional differences in the association between crime and poverty cannot be justified based on different levels of inequality or unemployment, which the literature correlates to property crimes. The variation in the observed rates of crime-poverty bias for different geographic locations could be influenced by cultural factors and the tendency to overestimate the equality of opportunities and social mobility in specific countries. These results have consequences for policy-making and open a new path of research for poverty mitigation with the focus not only on the poor but on society as a whole. Acting on the collective bias against the poor would facilitate the approval of poverty reduction policies, as well as the restoration of the dignity of the persons affected.

\end{abstract}

Artificial Intelligence (AI) provides insights that can trigger innovative interventions towards the UN Sustainable Development Goals (SDGs) \cite{Vinuesa2020a}. The \#1 UN SDG is the ``end of poverty in all its forms everywhere'', and there is an urgent call to find alternative paths to fight against poverty. The trend of global poverty reduction has been decreasing in the last decades  and the Covid-19 pandemic has erased the last four years of poverty mitigation. Rising inflation and the impact of the war in Ukraine derail the process of poverty mitigation even further \cite{PIP2023}. Worldwide, the World Bank estimates that 685 M people are living below the US\$2.15 a day poverty line \cite{PIP2023}. Poverty is a worldwide problem that affects not only the population in developing regions but also a significant percentage of people living in countries with thriving economies: in the United States, 11.6\% of the population (US 37.9 M people) are in a situation of poverty and 18.5 M live in extreme poverty \cite{UnitedNations2018}. 

The term \textit{undeserving poor} describes the difficulty for policy makers to approve and implement poverty reduction policies when the poor are blamed for their situation, since these policies are not popular \cite{Nunn2009}. Therefore, the blamefulness of the poor could have an impact on the actual poverty levels. Curto et al. \cite{Curto2022} provided evidence of bias against the poor, or aporophobia \cite{Cortina2017}, in Google, Twitter and Wikipedia word embeddings. This Natural Language Processing approach in Machine Learning has been widely used to identify biases within AI by representing words as vectors and measuring meaningful semantic relationships among them. It also allows us to reflect on societal biases, since the historical data used in AI has been produced by humans in real world scenarios.  

Caricatured narratives of the rich as being industrious and entrepreneurial while the poor are seen as wasters and scammers are at the root of this bias, which has been described as particularly prevalent in the United States \cite{UnitedNations2018}. The \textit{criminalization of poverty} refers to a discriminatory phenomenon where the poor are both associated with criminality and, at the same time, are being punished for being poor, generating a vicious circle. The criminal offenses devised for sleeping rough in many cities of the so-called developed countries are an example of a legalized punishment for being poor, since they affect the homeless directly. 

It must be highlighted that, to date, no evidence has been found that sustains the high level association between poverty and overall criminality rates. This correlation is complex to establish due to the multidimensional nature of poverty, the diversity of crimes (including violent and not violent), the method to identify them (self-reported and official reports), the potential police bias in the official crime rates reports, the different indicators of poverty used to study the potential correlation, the ages of the population in the sample, and geographic scope \cite{Thornberry1982}.   
However, recent studies report that factors such as a high poverty headcount ratio, high income inequality, and unemployment could have an impact on specific types of crimes, namely related to property \cite{Imran2018,Anser2020}. 

While the criminalization of poverty has been widely denounced as an instance of bias against the poor, or aporophobia, empirical evidence of this collective prejudice was missing in the literature. This brief report aims to fill in this gap. By investigating how poor people are viewed in society through the analysis of social media texts, namely tweets, we discover that poor and homeless people are often discussed in association with criminality. We devise a metric, crime-poverty-bias (CPB), as the difference between the percentage of utterances mentioning criminality and the poor and the percentage of utterances mentioning criminality and the rich, and compare CPB in Twitter for eight English-speaking counties. We present the CPB results per country together with the factors which could influence the increase of property criminality rate according to the literature, namely poverty, unemployment and inequality rates. The purpose of the study is to provide empirical evidence on the criminalization of poverty and shed some light on the reasons behind it. We aim to answer the following questions: Can the differences in the association between poverty and criminality in different countries be justified based on  the respective indicators of poverty, inequality, unemployment and criminality? Or are these differences due to a crime-poverty bias? The results have an impact on policy-making since CPB can hinder the acceptance of poverty mitigation policies by the public opinion.

\section*{Method}


We used the Twitter API to collect tweets in English from 25 August 2022 to 23 November 2022, pertaining to two groups: `poor' and `rich'. Since tweets can be up to 280 characters and include several sentences, we split each tweet into individual sentences. The corpus `poor' comprises tweet sentences that contain the terms 
\textit{the poor} (used as a noun as opposed to an adjective, as in `the poor performance'), \textit{poor people}, \textit{poor ppl}, \textit{poor folks}, \textit{poor families}, \textit{homeless}, \textit{on welfare}, \textit{welfare recipients}, \textit{low-income}, \textit{underprivileged}, \textit{disadvantaged}, \textit{lower class}. 
We excluded explicitly offensive terms that tend to be used in personal insults, such as \textit{trailer trash} or \textit{scrounger}. We also collected tweets related to the group `rich', using query terms 
\textit{the rich} (used as a noun), \textit{rich people}, \textit{rich ppl}, \textit{rich kids}, \textit{rich men}, \textit{rich folks}, \textit{rich guys}, \textit{rich elites}, \textit{rich families}, \textit{wealthy}, \textit{well-off}, \textit{upper-class}, \textit{upper class}, \textit{millionaires}, \textit{billionaires}, \textit{elite class}, \textit{privileged}, \textit{executives}. 
The single words \textit{poor} and \textit{rich} were not included as query terms because of their polysemy (they can apply to people, but can also be used to describe other things, e.g., `poor results', `rich dessert'). In total, there are over 1.3 million sentences in the corpus `poor' and over 1.9 million sentences in the corpus `rich'.

We were also interested in the geographical locations from which tweets originated. 
Unfortunately, only about 2\% of tweets included the exact geographical information. 
Therefore, in addition to the tweet location we relied on user location that tweeters voluntarily provide in their Twitter accounts, which was available for about 60\% of tweets. 
We automatically parsed user location descriptions to extract country information for the most frequently mentioned countries. 
Table~\ref{tab:number-tweets} shows the number of sentences in both corpora per geographical location.
In the following analysis, we focus on eight 
geographically-diverse English-speaking countries, for which both corpora contain at least 1,000 sentences each:  
the United States of America, the United Kingdom, Canada, India, Nigeria, Australia, South Africa, and Kenya.

To explore the themes commonly discussed with regard to poor people, we analyzed the content of sentences within the corpus `poor’ using an unsupervised topic modeling algorithm, BERTopic \cite{Grootendorst2022}. Topic modeling is a Machine Learning technique that aims to group texts semantically. As a first step, BERTopic converts texts to 384-dimensional vector representations so that semantically similar texts have similar representations. 
This conversion is performed using a sentence transformer, a large language model trained on over one billion sentences scraped from the web. 
Then, the vectors are clustered together using a density-based clustering technique HDBSCAN \citep{campello2013density}. 
The clustering algorithm identifies dense groups of semantically similar texts and leaves texts that do not fit any clusters as outliers. 
For computational efficiency, BERTopic was applied on a random sample of 600,000 sentences from the corpus `poor'. 


\begin{table}[tbhp]
\centering
\small 
\caption{The number of tweet sentences in the `poor' and `rich' corpora per geographical location.}
\begin{tabular}{lrr}
Location & `Poor' corpus & `Rich' corpus \\
\midrule
United States	& 326,993	& 460,848\\
United Kingdom	& 80,947	& 135,211\\
Canada	& 32,978	& 43,686\\
India	& 14,029	& 16,296\\
Nigeria	& 10,529	& 26,693\\
Australia	& 9,698	& 14,654\\
South Africa	& 7,729	& 8,600\\
Kenya	& 3,378	& 6,478\\[4pt]
Other locations & 337,252 & 461,437\\
No location information & 539,365 & 748,740\\[4pt]
\midrule
Total & 1,362,898 & 1,922,643\\
\bottomrule
\label{tab:number-tweets}
\end{tabular}

\end{table}

\begin{table}[h!]
\centering
\small
\caption{Example tweets relating criminality and poor people. The tweet texts were paraphrased to protect the privacy of the users.}
\begin{tabular}{p{7.5cm}}
\midrule
\textit{\textbf{Explicit statements associating the poor with criminality:}}\\
Put the homeless in jail and make work camps.\\
More and more homeless people are doing crime.\\
Homeless people in that area, criminals on the streets!!\\
[4pt]
\textit{\textbf{Statements opposing the criminalization of poverty, which elicit the underlying stereotype:}}\\
It's quite bold of you to claim that all homeless people are criminals.\\
\hangindent=2em So if you are in poverty you commit violent crimes and murder because you are disadvantaged?\\
\hangindent=2em Law enforcement and prisons are routinely used against poor people not for the reasons of safety, but to protect the wealthy and privileged.\\
\bottomrule
\label{tab:example-tweets}
\end{tabular}
\end{table}

\section*{Results and Discussion}

The topic modeling on sentences related to the group `poor' resulted in 142 extracted topics. 
Among the top topics in terms of frequency we find expected discussions such as the lack of affordable housing and (un)fair distribution of taxes among the socio-economic classes. We also discover themes relating to drug use, alcohol addiction and mental health issues associated with poverty. In this paper, we focus on the prominent topic related to criminality, which includes about 6,000 tweets. 
This topic is characterized by the presence of words such as \textit{crime}, \textit{police}, \textit{cops}, \textit{criminals}, and \textit{jail}. 
Some utterances explicitly associate poverty with crime, while others oppose such positions and criticize the systemic discrimination of the poor, including over-policing and disproportionate incarceration of poor and homeless people. However, the negation of stereotypes through counter-speech (e.g., ``not all homeless people are criminals'') is also a proof that these stereotypes exist in society \cite{Beukeboom2019}. 
Table~\ref{tab:example-tweets} shows some example tweets blaming the poor or denouncing social bias against the group. 

Since topic modelling techniques are typically used for qualitative studies, a quantitative analysis was carried out to complement the results obtained through BERTopic. We quantified 
 the percentage of sentences from the `poor' corpus (1.3M sentences) that contain the terms related to criminality, per country (Table~\ref{tab:rates-criminality-country}, row 1). 
The terms related to criminality include \textit{crime, crimes, criminal, criminals, criminalizing, jail, prison, arrest, arrested}. 
For comparison, we include the percentage of sentences from the `rich' corpus (1.9M sentences) that contain terms related  to criminality (row 2). 
We refer to the difference between rows 1 and 2 as the \textit{Crime-Poverty-Bias (CPB)},
since it measures the rate at which the poor are related to criminality as compared to the rich.


\begin{table*}
\centering
\footnotesize
\caption{Percentage of sentences 
that include the terms related to criminality in the `poor' and `rich' corpora and the difference in frequency, which constitutes the Crime-Poverty-Bias (CPB).}
\begin{tabular}{lrrrrrrrr}
Percentage of sentences & USA & Canada & South Africa & Kenya & UK & Nigeria & Australia & India\\
\midrule
1. in the `poor' corpus & 3.4 & 2.1 & 2.1 & 1.6 & 1.1  & 0.8 & 1.0 & 0.7 \\
2. in the `rich' corpus & 1.2 & 0.9 & 1.0 & 0.9 & 0.6 & 0.4 & 0.7 & 0.6\\
3. Crime-Poverty-Bias (CPB) & 2.2 & 1.2 & 1.1 & 0.7 & 0.5 & 0.4  & 0.3 & 0.1\\
\bottomrule
\label{tab:rates-criminality-country}
\end{tabular}

\end{table*}

\begin{figure*}[tph!]
\centering
\includegraphics[width= \textwidth]{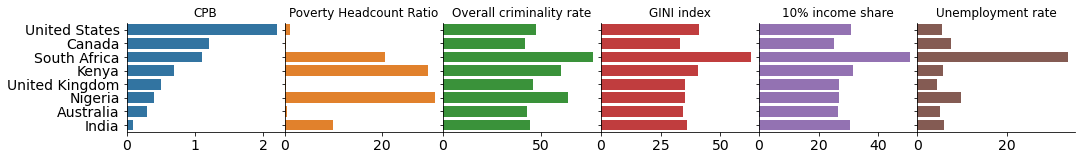}
\caption{For the countries in scope, overview of: the results obtained for Crime-Poverty-Bias (CPB), poverty headcount at \$ 2.5 a day (purchasing power adjusted prices), overall criminality rate, indicators of inequality (Gini index and 10\% income share) and unemployment rates. Sources: the CPB is obtained through the authors analysis of a corpus of tweets using the Natural Language Processing techniques. Poverty headcount ratio, Gini index, and 10\% income share rates (World Bank 2017 or nearest year). Unemployment rate (World Bank 2021) and overall criminality rate (worldpopulationreview.com).}\label{fig:poverty-economic}
\end{figure*}


The results shown in Table~\ref{tab:rates-criminality-country} indicate that CPB is highest in the United States, followed by Canada and South Africa.  Although the literature finds no correlation in general between overall crime rates and poverty, several factors have been identified that may potentially lead to an increase in property crime, such as income inequality and unemployment rate \cite{Anser2020,Imran2018}.  If the association between crime and poverty measured by the CPB reflects reality, we would expect higher CPB rates in countries rated higher on these measures. Therefore, to contextualize these outcomes, Figure~\ref{fig:poverty-economic} offers the CPB results together with each country's overall criminality rate, poverty headcount ratio at \$2.5 a day (purchasing power adjusted prices), inequality indicators (Gini Index and 10\% income share) and unemployment rate. However, we do not observe correlation between these indicators and the CPB for the eight countries. The United States has the highest CPB rates despite having lower poverty, criminality, inequality and unemployment rates than South Africa. Therefore, we must look to other causes for the overestimated correlation between poverty and criminality in the public opinion in the United States. The relatively low CPB results obtained for other countries such as Kenya, with higher poverty headcount and similar levels of inequality and unemployment rates to the United States, would support this hypothesis.

A potential explanation could be found on the narrative shared by the United States and Canada of being the ``land of opportunity'' where the rich and the poor are thought to have equal chance of success and an illusory emphasis on employment influences the discussion on the public social spending \cite{UnitedNations2018}. However, the principle of equal opportunity can be considered an oxymoron since every person is exposed to different opportunities in life from the moment of birth \cite{Sandel2020} and the job market for individuals with low educational qualifications, disability 
and with no assistance to find employment is very limited. The indicators of social mobility and inequality support the claim from the United Nations that the poor in the United States are overwhelmingly those born into poverty \cite{UnitedNations2018}: intergenerational social mobility in the United States from the bottom to the top income quintile is as low as 7.8\%, below European countries such as the UK, France, Italy, or Sweden \cite{Alesina2018}. In fact, intergenerational mobility has declined substantially over the last 150 years in the United States \cite{Song2020} and income inequality has been growing since the 1980s \cite{PIP2023}.

Although the use of Twitter is not representative of the total population within the countries in scope, the data obtained provides a first approach to measuring the phenomenon of the criminalization of the poor, which constitutes an instance of aporophobia. These preliminary results have an impact for poverty reduction policy making, because when the poor are considered ``undeserving of help'' it is more difficult for governments to approve laws to mitigate poverty. It is also harder for the people in need to overcome their situation when they are blamed for it and 
lack support from their community.

This brief report aims to initiate a new path of research for poverty mitigation, where the focus is not only on the redistribution of wealth but also on the mitigation of social bias against the poor. While the phenomenon of bias in terms of gender and race has been extensively analysed, bias against the poor has not received the attention it deserves both in AI and social sciences literature, despite the potential impact on the first global challenge identified by the United Nations.

\clearpage 

\bibliography{pnas-sample}
\bibliographystyle{acl_natbib}

\appendix

\section{Data Collection and Pre-processing}
\label{sec:appendix_tweets}

\subsection{Tweet Collection} 

We used the Twitter API to collect tweets in English from 25 August 2022 to 23 November 2022, pertaining to two groups: `poor' and `rich'. 
The initial set of query terms has been assembled from social psychology literature, and expanded with synonyms and related terms. 
Then, a one-week sample of tweets collected using this initial set has been manually examined, and terms that resulted in very small numbers of retrieved tweets or in many irrelevant tweets were discarded. 
The final list of query terms for the group 'poor' included: \textit{the poor} (used as a noun as opposed to an adjective as in `the poor performance'), \textit{poor people}, \textit{poor ppl}, \textit{poor folks}, \textit{poor families}, \textit{homeless}, \textit{on welfare}, \textit{welfare recipients}, \textit{low-income}, \textit{underprivileged}, \textit{disadvantaged}, \textit{lower class}. 
We excluded explicitly offensive terms that tend to be used in personal insults, such as \textit{trailer trash} or \textit{scrounger}. For the group `rich' we used the following query terms: \textit{the rich} (used as a noun), \textit{rich people}, \textit{rich ppl}, \textit{rich kids}, \textit{rich men}, \textit{rich folks}, \textit{rich guys}, \textit{rich elites}, \textit{rich families}, \textit{wealthy}, \textit{well-off}, \textit{upper-class}, \textit{upper class}, \textit{millionaires}, \textit{billionaires}, \textit{elite class}, \textit{privileged}, \textit{executives}. 
The single words \textit{poor} and \textit{rich} were not included as query terms because of their polysemy (they can apply to people, but can also be used to describe other things, e.g., `poor results', `rich dessert'). 

\subsection{Data Pre-processing} 

We filtered out re-tweets, tweets with URLs to external websites, tweets with more than five hashtags, and tweets from user accounts that have the word \textit{bot} in their user or screen names. 
This step helped remove advertisements, spam, news headlines, and other non-personal communications. 
Further, tweets containing query terms from both `poor' and `rich' groups were also excluded. 
The remaining tweets were split into individual sentences and only sentences that included at least one of the query terms were kept. 
User mentions have been replaced with `@user' and query terms have been masked with `<target>' to reduce the bias from the query terms in the analysis.
In total, there are over 1.3 million sentences in the corpus `poor' and over 1.9 million sentences in the corpus `rich'.

\subsection{User Location Identification} 

To identify the location from which a tweet originated, we used both tweet location and user location fields that Twitter provides. Only about 2\% of the tweets included the exact geographical information from which the tweet was sent (the field `place').  
User location was specified in about 60\% of the tweets. 
This information was presented as a free-form text, and tweeters were often very creative in describing their location (e.g., ``somewhere on Earth''). 
We automatically parsed user location descriptions to extract country information for the most frequently mentioned countries. 
In the absence of a country name, we considered the mentions of U.S. states, Canadian provinces, and major cities in the U.S., U.K. and Canada, since these were also commonly used by tweeters. (Major cities from other countries were rarely used without the country name.)

\section{Topic Modeling}
\label{sec:appendix_topic}

BERTopic \citep{Grootendorst2022} is a flexible state-of-the-art toolkit for unsupervised, semi-supervised, and supervised topic modeling. 
The input documents are first converted to a numerical vector (called embedding) space using techniques such as sentence transformers. Then the dimensionality of the embedding space is reduced with techniques like PCA or UMAP since the clustering methods usually are more effective in low-dimensional spaces.  
The core component of BERTopic is a density-based clustering technique HDBSCAN \citep{campello2013density}, which can produce clusters of different shapes and leave documents that do not fit any clusters as outliers. This suited our case well as we wanted to discover the most commonly discussed topics in tweets mentioning poor people. 
The discovered topics are then represented with topic words, which are identified using class-based TF-IDF (c-TF-IDF). Topic words are the words that tend to frequently appear in the topic of interest and less frequently in the other topics. 

We applied BERTopic in the unsupervised mode using the following settings and parameters. For converting text to numerical representations, we used the sentence transformers\footnote{\url{https://github.com/UKPLab/sentence-transformers}} method based on the \textit{all-MiniLM-L6-v2}\footnote{\url{https://www.sbert.net/docs/pretrained_models.html}} pre-trained embedding model. For the vectorizer model, we used the CountVectorizer method,\footnote{\url{https://scikit-learn.org/stable/modules/generated/sklearn.feature_extraction.text.CountVectorizer.html}} and 
removed English stopwords and 
terms that appear in less than 5\% of the sentences ($min\_df = 0.05$). 
For the HDBSCAN clustering algorithm, we specified the minimum size of the clusters as $min\_cluster\_size=500$. For all the other parameters, the default settings of the BERTopic package were used. 

\end{document}